\documentclass{llncs}
\usepackage{graphicx, array, makeidx, amsmath, amssymb, url}

\usepackage{amsthm}
\usepackage[usenames,dvipsnames]{color}
\usepackage[autolanguage]{numprint}

\begin{document}
\hyphenation{con-fi-gu-ra-tion}
\frontmatter			
\pagestyle{headings}	
\mainmatter			

\title{Automatic Observer Script\\ for StarCraft: Brood War Bot Games\\ (technical report)}
\titlerunning{SSCAIT Auto-Observer}  

\author{Bj\"{o}rn Persson Mattsson$^1$ \and Tom\'{a}\v{s} Vajda$^2$ \and Michal \v{C}ertick\'{y}$^3$}
\authorrunning{Mattsson, Vajda, \v{C}ertick\'{y}} 

\theoremstyle{definition}
\newtheorem{exmp}{Example}


\urldef{\mailsa}\path|bjorn.pm@plankter.se|
\urldef{\mailsb}\path|thomas.vajda@gmail.com|
\urldef{\mailsc}\path|certicky@agents.fel.cvut.cz|

\institute{
\mailsa\\ \mailsb\\ \mailsc
\vskip 5pt
$^1$Chalmers University of Technology, Sweden\\
$^2$Comenius University, Bratislava, Slovakia\\
$^3$Czech Technical University in Prague, Czech Republic\\
}

\maketitle              

\begin{abstract}
This short report describes an automated BWAPI-based script developed for live streams of a StarCraft Brood War bot tournament, SSCAIT. The script controls the in-game camera in order to follow the relevant events and improve the viewer experience. We enumerate its novel features and provide a few implementation notes.
\end{abstract}
\section{Introduction}
Since 2010, multiple competitions for StarCraft: Brood War BWAPI\footnote{\url{http://github.com/bwapi/bwapi}} bots have been founded, including long-term {\em SSCAIT: Student StarCraft AI Tournament}\footnote{\url{http://sscaitournament.com/}} and {\em Brood War Bots Ladder}\footnote{\url{http://bots-stats.krasi0.com/}} as well as short-term events organized as parts of research conferences {\em AIIDE} and {\em CIG}. The growing community behind StarCraft AI research and increasing interest of the public has lead the organizers of the long-term tournaments to stream the games publicly. The need to stream the bot games 24 hours a day raises two automation-related challenges:
\begin{enumerate}
  \item The games need to be started, ended and evaluated automatically.
  \item The in-game camera needs to automatically follow all the interesting events to make the stream attractive for the viewers.
\end{enumerate}
To solve the first challenge, one can simply use the open-source Tournament Manager Software\footnote{\url{http://webdocs.cs.ualberta.ca/~cdavid/starcraftaicomp/tm.shtml}} developed at University of Alberta (SSCAIT tournament uses a modification of this tool).
For the second challenge, however, this tool has proven to be insufficient. Even though it comes with an auto-observer module capable of moving the in-game camera around, this module has proven to be too simplistic. After numerous complaints by the stream viewers, we decided to implement a new observer module for SSCAIT tournament.

\section{SSCAIT Observer}
The SSCAIT observer is implemented in the C++ framework BWAPI and compiled as a part of the ``tournament module'' DLL (which provides additional, camera-unrelated functionality).   
It runs an automated spectating behaviour that is based primarily on game events, priorities and timers. It accomplishes this by setting the game screen position to the in-game location that should be shown. For the remainder of the text, this behaviour will be termed as ``moving the camera''.

The camera will always try to focus on game events that are deemed important for the viewers (e.g. units attacking each other, the creation of new units, and so on). If several events happen simultaneously, the camera uses a set of predefined priorities to determine which event to focus on. In order to improve the viewing experience, timers are used to make sure that the camera does not move too often.

\subsection{Time Limits and Priorities}
Two time limits are used by the SSCAIT observer: one defining the minimum time ($t_{min}$) that must pass before the camera is allowed to focus on a higher prioritized event, and another defining the time ($t_{max}$) that must pass before the camera will focus on any new event (even one with lower priority).

\begin{exmp}
Let us say that $t_{min} = 50$ and $t_{max} = 150$ (the time values are specified in logical game frames). Here is an example of how this mechanism works:
\begin{itemize}
\item At time $t = 0$, event $e_1$ with priority 1 occurs.
	$\rightarrow$ camera focuses on event $e_1$.
\item At time $t = 40$, event $e_2$ with priority 2 occurs.
	$\rightarrow$ since less time than $t_{min}$ has passed since the last event, the camera remains focused on $e_1$.
\item At time $t = 60$, event $e_3$ with priority 0 occurs.
	$\rightarrow$ since $e_3$ has lower priority than the current focus $e_1$ and less time than $t_{max}$ has passed, the camera remains focused on $e_1$.
\item At time $t = 160$, event $e_4$ with priority 0 occurs.
	$\rightarrow$ more time than $t_{max}$ has passed since the last focus change, so the camera focuses on event $e_4$.
\item At time $t = 220$, event $e_5$ with priority 3 occurs.
	$\rightarrow$ $e_5$ has higher priority than the current focus $e_4$ and more time than $t_{min}$ has passed, so the camera focuses on event $e_5$.
\end{itemize}
\end{exmp}

\subsection{Smooth Camera Movement}
When the camera of the SSCAIT observer focuses on an event, it does not simply ``teleport'' the camera to the focus position. Instead, it always moves the camera smoothly towards the desired location by the simple update rule:
$$
pos_{n+1} = m \cdot (fPos - pos_{n})
$$
where $pos$ is the current position of the camera, $fPos$ is the desired focus position, and $m$ is a movement factor between $0$ and $1$. A movement factor of $m = 0.1$ would mean that the camera moves $10\%$ of the distance between the current camera position and the focus position. The SSCAIT observer currently uses $m = 0.1$ and updates the camera position on every game frame.

Another feature of the SSCAIT observer is that the camera not only can focus on a position. It can also focus on a unit. This functionality means that the camera can follow units (e.g. an army or a scouting worker) around the map instead of remaining still at a position that the relevant unit might have moved away from.

\subsection{Events}
The types of game events that the camera focuses on and the priorities currently associated with each of them are:
\begin{itemize}
\item A unit is under attack. Priority: 3
\item A unit is performing an attack. Priority: 3
\item A worker is scouting.
	\begin{itemize}
	  \item[-] If scout worker is close to a potential enemy base, priority: 2
	  \item[-] Otherwise, priority: 0
	\end{itemize} 
	{\em Note:} A worker is only counted as a scout if it is not in its own main base and if the frame count is less than 7500 (approximately 8 minutes in terms of in-game time).
\item A drop is performed. Priority: 2\\
	{\em Note:} Here, a drop is counted as a non-empty transport unit that is close to a potential enemy base.
\item A group of army units are positioned closely together. Priority: 1\\
	{\em Note:} A definition of ``army unit'' can be found below.
\item A unit is created. Priority: 1
\end{itemize}

Some of these events are detected by looping over all accessible units and checking if the condition is fulfilled, while others are purely event-based, in the sense that the game notifies the observer module as the event happens.

\subsection{Groups of Army Units}
In order to detect armies in a good way, a definition of ``army unit'' needed to be formed. Since a group of workers gathering resources should not be classified as an ``army'', the observer excludes workers when searching for army units. Other unit types that are not counted as army units are structures, larvae, overlords and spider mines. 

The reason to exclude overlords and spider mines might need an explanation: some of the Zerg bots playing in the SSCAIT tournament tend to clump overlords together in a corner of the map, while some Terran bots use the strategy of creating very dense minefields. Most viewers would probably agree that watching a minefield for large parts of a match is not the most thrilling experience.

\subsection{Dynamic screen resolution}
When StarCraft was originally released in 1998, it used a constant screen resolution of $640\times 480$ px. Modern displays have much higher resolutions and different aspect ratios. Consequently, StarCraft is only able to make use of a small part of the screen. Fortunatelly, it is possible to use various ``resolution hacks" to force the game to run on higher screen resolutions, displaying a larger part of the map. Even though this is considered cheating by human players, bots are not affected by the field of view in any way and gain no advantage from higher resolutions. At the same time, increasing the screen size makes up for more attractive and modern looking streams. The SSCAIT observer is compatible with resolution hacks and provides a support for any screen resolution (specified in the configuration file). Such compatibility is a simple matter of dynamically determining the screen center and showing the events there. 

\section{Conclusion}
Thanks to the custom observer script  described in this text, the SSCAIT tournament currently broadcasts a visually attractive live stream of all its bot games at high-definition, widescreen resolution. It selects the most interesting in-game situations, and transitions between them smoothly to improve the viewer experience. The stream can be watched at the tournament's website\footnote{\url{http://sscaitournament.com/}}.  

We are planning to release the script as an open-source software in the near future. In the meantime, it is available on-demand.

%
%

\clearpage
\addtocmark[2]{Author Index} 
\renewcommand{\indexname}{Author Index}
\printindex
\clearpage
\addtocmark[2]{Subject Index} 
\markboth{Subject Index}{Subject Index}
\renewcommand{\indexname}{Subject Index}
\end{document}